%% file: 0root.tex
\documentclass[letterpaper, 10 pt, conference]{ieeeconf}  

\IEEEoverridecommandlockouts                              

\usepackage{graphics} 
\usepackage[dvipdfmx]{graphicx}
\usepackage{mathptmx} 
\usepackage{times} 
\usepackage{amsmath} 
\usepackage{amssymb}  
\usepackage{algorithm,algorithmic}
\usepackage{subcaption}
\title{\LARGE \bf
lamBERT: Language and Action Learning Using Multimodal BERT
}

\author{Kazuki Miyazawa$^{1}$, Tatsuya Aoki$^{1,2}$, Takato Horii$^{1}$, and Takayuki Nagai$^{1,3}$
\thanks{This research was supported by JST CREST (JPMJCR15E3) and JSPS KAKENHI Grant Number JP19J23364.
}
\thanks{$^{1}$Graduate School of Engineering Science, Osaka University, Osaka, Japan
        {\tt\footnotesize\{k.miyazawa, t.aoki\}@rlg.sys.es.osaka-u.ac.jp, 
        \{takato, nagai\}@sys.es.osaka-u.ac.jp }}%
\thanks{$^{2}$Graduate School of Informatics and Engineering, The University of Electro-Communications, Tokyo, Japan}%
\thanks{$^{3}$Artificial Intelligence Exploration Research Center, The University of Electro-Communications, Tokyo, Japan}%
}

\begin{document}

\maketitle
\thispagestyle{empty}
\pagestyle{empty}

\begin{abstract}

Recently, the bidirectional encoder representations from transformers (BERT) model has attracted much attention in the field of natural language processing, owing to its high performance in language understanding-related tasks.
The BERT model learns language representation that can be adapted to various tasks via pre-training using a large corpus in an unsupervised manner.
This study proposes the language and action learning using multimodal BERT (lamBERT) model that enables the learning of language and actions by 1) extending the BERT model to multimodal representation and 2) integrating it with reinforcement learning.
To verify the proposed model, an experiment is conducted in a grid environment that requires language understanding for the agent to act properly.
As a result, the lamBERT model obtained higher rewards in multitask settings and transfer settings when compared to other models, such as the convolutional neural network-based model and the lamBERT model without pre-training.

\end{abstract}

\input{1intro} 
\input{2prop} 
\input{3experiments} 
\input{4discussion} 
\input{5conclusion} 

\addtolength{\textheight}{-12cm}   


\bibliographystyle{ieeetr}
\bibliography{reference}

\end{document}

%% file: 1intro.tex
\section{Introduction}

Human support robots, that can work together with us and support our lives in our surrounding environments, are expected to be realized in the near future. 
However, despite the rapid development of AI and robotics in recent years, such robots have not yet been realized. 
One of the reasons is that it is currently difficult for such robots to ``understand'' the physical environment and human language to ``act'' appropriately. 
As human language commands are ambiguous, and the situation changes sequentially, the robot is required to keep adapting to such a change while acquiring knowledge based on its own experience in a bottom-up manner. 
Therefore, it is not practicable to collect a huge amount of labeled data in advance.
In other words, it is necessary for the robot to learn the representation and actions based on data collected through its own actions.

To completely understand complex environments, multiple modalities should be handled.
The relationship between the modalities is the key to understanding the complex physical world.
Research has been conducted on real-world understanding including language acquisition by robots \cite{taniguchi2016symbol}.
In this context, ``understanding'' is defined as the prediction of unobserved information through concepts. 
The concept is a structure of multimodal information obtained via multiple sensors of the robot. 
By predicting the information through the concept, the robot can predict unobserved information more quickly and accurately. 
This is because detailed information including noise is discarded in the concept formation process, and the important information is collected. 
In \cite{nakamura2009grounding}, the authors have proposed a model that categorizes the multimodal information obtained by real robots using a probabilistic generative model to form concepts and to predict the unobserved information. 
Moreover, this framework has been extended to a model that learns the concepts and the language simultaneously \cite{araki2012online,araki2013long}. 
Despite these achievements, there still exist some problems in the multimodal representation learning using these probabilistic generative models. 
First, it is not clear whether the knowledge required for the robot operating in complex environments can be represented in the form of generative processes. 
Second, the inference via latent variables can learn the multimodal representation according to the specific task (classification and reinforcement learning, etc.); however, it might be difficult to acquire effective representation for multiple tasks. 
In the probabilistic generative models, the relationship between the latent variables is provided in the form of a directed acyclic graph (DAG). 
The drawback is that it is unknown as to how the structural design of the DAG affects the performance of the task. 
In any case, these factors can pose problems to the robot's operation in complex environments. 
Although the multimodal representation learning has been studied for deep generative models such as the variational autoencoders (VAEs) \cite{suzuki2016joint,NIPS2019_9702}, they also pose similar problems. 
It is unclear as to which mechanism is suitable for learning the multimodal representations at all.

Recently, deep reinforcement learning has greatly contributed to the development of robot action learning. 
Various models have been proposed so far, and their usefulness has been demonstrated in real robots \cite{levine2016end}. 
Moreover, transferring the experience learned from a certain environment to a new environment is very important in the real robot learning scenario, where the available samples are limited. 
To achieve this, several methods have been proposed in which meta-knowledge, that can be used across multiple tasks, is acquired and used to adapt to a new environment \cite{finn2017model}. 
Furthermore, attempts are being made to learn knowledge applicable to multiple robots from a large amount of data acquired by the robots \cite{dasari2019robonet}. 
Thus, the problem of acquiring knowledge across multiple tasks and making appropriate decisions in a new environment is crucial, but not completely solved thus far. 

On the contrary, natural language processing (NLP) in recent years has begun to fare better than humans in tasks that require language understanding, and it has therefore received considerable attention.
These advances are based on the language models, such as ELMo \cite{peters2018deep} and BERT \cite{devlin2018bert}, that perform unsupervised pre-training using a large text corpus. 
In particular, the BERT model enables the pre-training for various NLP tasks. 
The BERT model is now considered as a general pre-training model for language understanding. 
The pre-training is based solely on self-supervised learning, such as the ``mask prediction task'' and ``next sentence prediction task,'' which requires no labels for the training data. 
The pre-trained BERT model scored the highest among all models on eight different NLP tasks.

Multimodal representation learning that can be applied to multiple tasks is also a critical aspect to be considered regarding robots.
One of the recent developments concerning BERT is its extension to multimodal data. 
Models that enable multimodal representation learning using the structure and learning mechanism of BERT have been proposed \cite{sun2019videobert, lu2019vilbert, qi2020imagebert}. 
By further expanding these frameworks, we believe it is possible for robots to understand the environment wherein they can respond to human language by connecting it to the real world, which is referred to as ``symbol grounding.'' 
In addition, as the robot can act autonomously, the robot can continuously obtain information corresponding to the pre-training data in NLP tasks via its own actions and keep updating the knowledge of the world and language.

In this study, we propose lamBERT (language and action learning using multimodal BERT), in which the robot learns concepts from multiple sensory inputs using the multimodal BERT structure, and  learns actions using the learned multimodal representation and reinforcement learning algorithm. 
We verify the usefulness of the model using a grid environment that requires language instructions to be understood and the structure and learning method of the BERT model to be shown as effective for multitask learning and transfer learning. 

\section{Related work}
Related work can be broadly classified into three viewpoints, each of which is described below. 
\subsection{Multimodal representation learning}
For a robot to understand the real world, organization of multimodal information and prediction of unobserved information are inevitable processes. 
Various models for performing the multimodal representation learning that enable such predictions have been proposed, as mentioned earlier. 
For example, JMVAE \cite{suzuki2016joint} and TELBO \cite{vedantam2017generative} use VAEs to acquire a multimodal representation through latent variable $z$ and enable the mutual prediction of information for each modality. 
In VideoBERT \cite{sun2019videobert}, ViLBERT \cite{lu2019vilbert}, and ImageBERT \cite{qi2020imagebert}, the multimodal representation of language and images is obtained using the structure of BERT, and cross-modal mutual prediction is made possible. 
The multimodal latent Dirichlet allocation (MLDAs) \cite{nakamura2009grounding} and spatial concept SLAM (SpCoSLAM) \cite{SPCOSLAM2017IROS} are examples of the multimodal representation learning using probabilistic generative models. 
By acquiring and structuring the multimodal information, these models enable real robots to learn object and spatial concepts and predict unobserved information. 

\subsection{Transfer learning}
Various models have been proposed for learning and acting while exhibiting adaptation to multiple environments. 
UNREAL \cite{jaderberg2016reinforcement} shows that adding an auxiliary task to a reinforcement learning task improves the performance in the environment to which it is transferred. 
Similarly, the performance of the autonomous driving cars can be improved by adding certain tasks, such as prediction of depth and segmented images from the RGB image \cite{li2018rethinking}. 
Moreover, as collecting data for the robot in the real world is expensive, a method called domain randomization, wherein the training is conducted in the simulation following which the model is fine-tuned in the real world, has also attracted attention \cite{chebotar2019closing}.

\subsection{Integrated models} 
As previously mentioned, research focused solely on representation learning or reinforcement learning, as well as on models that handle multimodal representation learning and reinforcement learning in a unified manner, have been proposed. 
Hermann {\it et al.} proposed a model for learning agent actions and language in a three-dimensional maze environment \cite{hermann2017grounded}. 
We proposed a model, which we call the integrated cognitive architecture, for learning concepts, language, and actions using a robot in the real world \cite{miyazawa2019integrated}. 
The Visual-and-Language Navigation (VLN) is a task that needs to deal with the multimodal information and make decisions. 
The models that realize this task can be considered as integrated models. 
Zhu {\it et al.} have proposed a VLN agent model \cite{zhu2019vision} that uses attention to integrate the image and language information. 

In this study, to realize an autonomous agent that can act by understanding human language, we propose the lamBERT model that can learn language and actions using the structure and learning method of BERT employed in the NLP tasks.
This is realized by extending the BERT model to multimodal data and integrating the network of multimodal representation learning with deep reinforcement learning. 
The contributions of this paper are twofold: 1) BERT is shown to be effective for multimodal representation learning in a reinforcement learning scenario and 2) we demonstrate that BERT is suitable for the transfer reinforcement learning. 
To the best of our knowledge, this is the first attempt of applying BERT to reinforcement learning. 

%% file: 2prop.tex
\section{Proposed method}
In this section, we introduce BERT briefly followed by the explanations of the structure and learning method of the proposed lamBERT model. 
\subsection{Overview of BERT}
The BERT model is a pre-trained model used for solving NLP tasks. 
This model is trained using a large amount of text data in a self-supervised manner. 
Upon using this pre-training scheme, the BERT can adapt to various tasks by employing the fine-tuning. 
The BERT model consists of multilayered transformers that utilize a multi-head-self-attention mechanism. 
The input to the BERT model is the sum of the following three embedded vectors:
token embeddings that embed the language token information, 
position embeddings that embed the information indicating the order of words in the sequence, 
and segment embeddings that embed information for discriminating between two sentences. 
In the pre-training phase, the language model is trained by performing two unsupervised learning tasks on this input. 
One is the mask prediction task for predicting the masked tokens, which is called a masked language model (Masked LM). 
The other is the next sentence prediction task that determines whether two sentences in the input are connected. 
By performing fine-tuning for a specific task using the weights obtained by the pre-training, the BERT model has demonstrated high performance in various NLP tasks. 

\subsection{lamBERT model}
\begin{figure}[tb]
	\begin{center}
	   \includegraphics[width=1.0\linewidth]{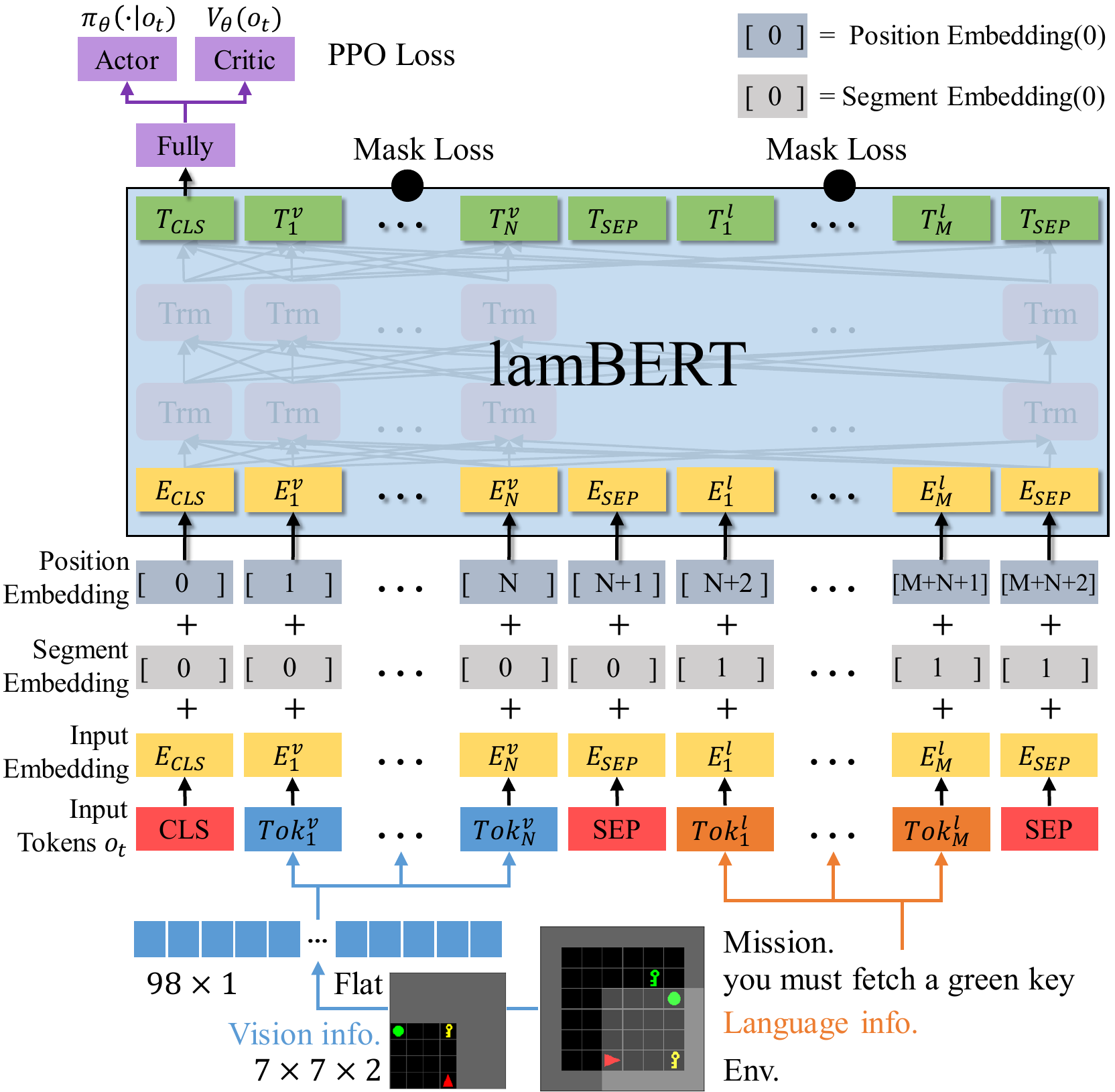}
	   \caption{
	   Overview of lamBERT: 
	   lamBERT outputs policies $\pi_{\theta}(\cdot|o_t)$ and values $V_{\theta}(o_t)$ using visual and language information as input tokens $o_t$. 
	   The visual information is input as a tokenized sequence of a flattened image $Tok_{n}^{v}$, and the language information is input as tokenized words $Tok_{m}^{l}$. 
	   Model learning is performed by inputting a masked token sequence and calculating the loss function.
      The sum of ``Proximal Policy Optimizations (PPO) Loss'' and ``Mask Loss'' is the loss of the entire model. 
      The policy is calculated using the input token without the masking at the time of action. 
	   }
	   \label{fig:lamBERT}
  \end{center}
\end{figure}

An overview of the lamBERT model is shown in Fig. \ref{fig:lamBERT}. 
The lamBERT model uses two modality tokens, which are vision tokens $Tok_{n}^{v}$ and language tokens $Tok_{m}^{l}$, as input $o_t$, learns the multimodal representation of these two modalities obtained through the BERT structure, and outputs the policies $\pi_{\theta}(\cdot|o_t)$ and values $V_{\theta}(o_t)$ via an actor-critic network.
The model parameter $\theta$ is updated by the reinforcement learning algorithm and the mask prediction task. 
We now explain the details of the lamBERT model.

\subsubsection{Multimodal representation}
The proposed model uses the BERT structure extended to multimodal input. 
This is realized using the BERT structure and the token sequence that combines visual information and language information as the input.
This method is inspired by the structure of the model to handle multimodal information with the BERT structure \cite{sun2019videobert}, \cite{lu2019vilbert}, \cite{kiela2019supervised}.
There are various ways to input the visual information to BERT, but in this model, the vision token sequence uses a flattened image directly. 
The proposed model assumes that the visual information and language information are input simultaneously.
In this research, we use the image observed by the agents in a grid environment as shown in Fig.  \ref{fig:lamBERT}. 

The input tokens are composed of the vision tokens and the language tokens by inserting the classification (CLS) token and separation (SEP) token, as shown in Fig. \ref{fig:lamBERT}.
If the sequence length of the input tokens is shorter than the maximum length, a padding is performed using a PAD token.
The position tokens and the segmentation tokens are used as the representation for the difference between the token position information and modality, respectively.
The position tokens assign a position ID from $0$ to the sequence length to the token string that combines the vision tokens and the language tokens.
The segmentation tokens assign $0$ to the vision tokens and $1$ to the language tokens.
The final input to the model is to map the input tokens, position tokens, and segmentation tokens into a $d$-dimensional embedding space using a learnable network, and then take the sum and use it as an input to the model.

To learn multimodal representation, the mask prediction task performed by the original BERT applies also to the proposed method.
The mask prediction task in the proposed model is performed as follows.
First, 5\% of the tokens are selected from input tokens (vision + language) as mask prediction candidates.
Then, 80\% of the mask prediction candidates are replaced with the ``MASK" token,
and 10\% of the mask prediction candidates are replaced with another random token.
No conversion of the last 10\% of mask prediction candidates should occur.
The masked token sequence $ o_{t \setminus mt} $ is input to the lamBERT, and the lamBERT model learns to generate the original token sequence.
The masked token sequences $ o_{t \setminus mt} $ is created $D$ kinds per one token sequence $ o_{t} $ and the model uses them as the training data.
For learning, the sum of the cross-entropy loss of only the tokens that have become mask candidates is used.
The loss function of the mask prediction task is expressed as
\begin{equation}
    L_{ml}=  \frac{1}{|C|} \sum_{cid}^{C}{ \left[ -{\rm log}\left(\frac{{\rm exp}(x_{cid})}{\sum_{id}^{ID} {\rm exp}(x_{id})}\right) \right] }, 
\end{equation}
where $x$, $ID$, and $ cid \in C$ represent the final embedding vector of the masked token, size of all token IDs, and ID of the token selected as the mask candidate token, respectively. 
$|C|$ represents the total number of the tokens selected as the mask candidate token.

In modalities such as visual information and language information, where the number of tokens differ between the modalities, the effect of the mask prediction task is dominated by modalities with many tokens.
Therefore, when selecting mask candidates, we adjust the probability of selection as a mask candidate.
When the length of each modality sequence is $N$ and $M$, respectively, each token is selected for mask prediction with the probabilities ${\rm P}(Tok_{n}^{v}=mask) = \frac{1}{2N}$ and ${\rm P}(Tok_{m}^{l}=mask) = \frac{1}{2M} $.


\subsubsection{Action learning}
We use Proximal Policy Optimizations (PPO) \cite{schulman2017proximal}, a type of actor-critic algorithm, for the action learning.
The actor and the critic are calculated using the embedding vector of the last layer of the CLS token as shown in Fig. \ref{fig:lamBERT}.
In this manner, the action decision becomes possible using the structure of BERT.
The PPO learning is performed based on the following formula using data sampled by the agents acting in the environment.

\begin{eqnarray}
  L_{ppo} &=& -{\rm min} \left [ r_t(\theta) \hat{A}_t, ~{\rm clip} \left ( r_t(\theta),1-\epsilon,1+\epsilon  \right ) \hat{A}_t\right ] \nonumber \\
  &+& c_1 L_{t}^{VF}(\theta) - c_2 S[\pi_\theta]( o_{t \setminus mt} ),
\end{eqnarray}
where, $r_t(\theta)$ is the probability ratio:
\begin{eqnarray}
  r_t(\theta) = \frac{\pi_\theta(a_t | o_{t \setminus mt} )}  {\pi_{\theta_{old}}( a_t | o_t )},
\end{eqnarray}
where $a_t$, $o_t$, $o_{t \setminus mt}$, and $\hat{A}_t$ represent the agent action, input tokens, masked input tokens, and estimated advantage, respectively. 
$L_{t}^{VF}$, $S$, and $\epsilon$ represent the squared-error loss for value function, entropy bonus, and hyperparameter for clipping, respectively. 
$c_1$ and $c_2$ are weights for the value and the entropy bonus, respectively. 
Note that the policy $\pi_{\theta_{old}}$ is calculated using input tokens $o_t$ during action, and the policy $\pi_{\theta}$ is calculated using masked input tokens $ o_{t \setminus mt} $ during the model update.
\subsubsection{Leaning}
In reinforcement learning, unlike the problem setting in NLP, it is rare that a large amount of environment data is available before learning in the environment.
It is necessary for the agent to acquire the data itself and acquire the necessary knowledge.
Therefore, unlike the BERT model used in NLP, the lamBERT model obtains the data used for mask prediction learning through trial and error in the environment. 
In other words, the lamBERT model does not have two steps of pre-training and fine-tuning as performed in the original BERT model.
The lamBERT model updates by learning the mask prediction task simultaneously with data collection and policy learning by reinforcement learning. 
The loss function of the lamBERT model is expressed as 
\begin{equation}
  L = L_{ml} + L_{ppo}.
\end{equation}
The learning flow of the lamBERT model is shown in Algorithm 1.

\begin{algorithm}[tb]
 \caption{lamBERT model Algorithm}
 \begin{algorithmic}[1]
  \FOR {$iteration = 1,2,...$}
  \FOR {$actor = 1,2,...,N$}
  \STATE Act by policy $\pi_{\theta_{old}}$ for $T$ timesteps
  \STATE Calculate advantage estimates $\hat{A}_1,...,\hat{A}_T$
  \ENDFOR
  \FOR {$duple = 1,2,...,D$}
  \STATE Generate $T \times N$ masked token sequence $ o_{ \setminus mt} $
  \ENDFOR
  \STATE Optimize $L$ with respect to $\theta$ with K epochs and minibatch size $B < N \times T \times D$
  \STATE $\theta_{old} \leftarrow \theta$
  \ENDFOR
 \label{algorithm:lamBERT}
 \end{algorithmic} 
 \end{algorithm}

\subsubsection{Transfer}
Transfer learning is performed by applying the model learned using the above algorithm in a certain environment to the other environments.
In a certain environment, the lamBERT model adapts to the environment by updating the model using the PPO algorithm, and learns general knowledge by applying mask prediction tasks to new languages and environments.

%% file: 3experiments.tex
\section{Experiments}
To verify the effectiveness of the proposed method, we examined action learning and language learning in a multitasking environment and the transfer of knowledge to a new environment for a grid environment with language instructions.
First, we describe the experimental environment, the experimental setting, and finally, describe the experimental results.

\subsubsection{Experimental environment}
\label{section:Experimental_Environment}
We used the gym minigrid \cite{chevalier-boisvert2018babyai} for the experimental environment.
In this environment, objects (key, ball, and box, etc.) and doors are placed in the grid environment.
The agent can take any of seven actions: straight ahead, left turn, right turn, pick up, drop an object, toggle an object, and done. 
A language instruction (mission) in a natural language is given as a task to be performed by an agent.
The agent gets a reward for taking action to achieve this instruction.

\begin{figure}[tb]
	\begin{center}
	   \includegraphics[width=1.0\linewidth]{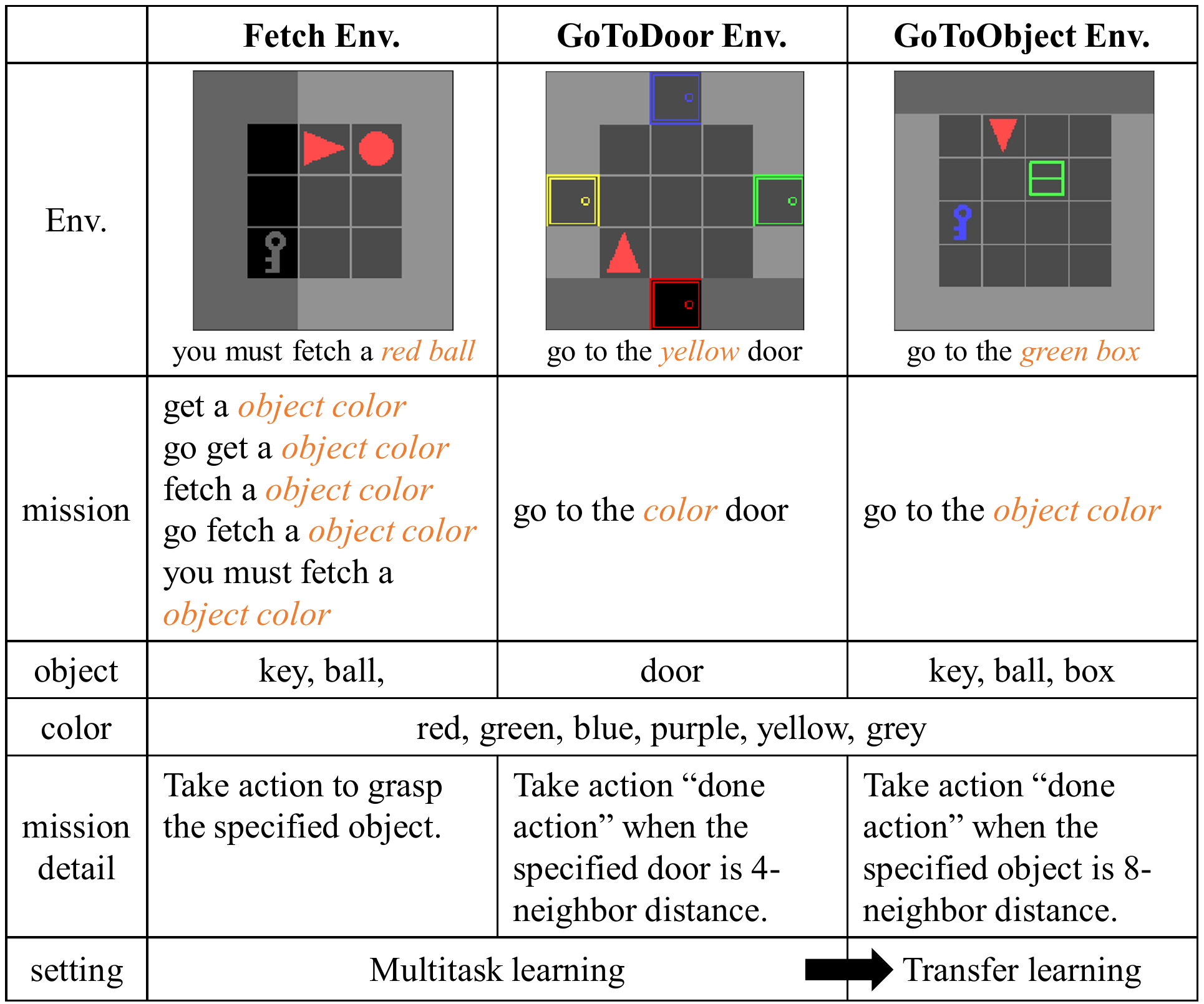}
	   \caption{
	   Experimental environments: 
	   Agents (shown as red triangle) receive rewards only when they achieve the mission.
	   The reward value is a maximum value of $1$ and decays over time. 
	   The color and object of the mission are randomly selected from those present in the environment.
	   For the fetch environment's mission, one fixed phrase is randomly selected from the above missions. 
	   The agent's and the initial object positions in each episode are given at random. 
	   The agent's visual information is the partial observation of $7\times7$ squares in front of the agent (the highlighted area). 
	   The visual information has two channels of discrete values representing type and color of the object. 
	   }
	   \label{fig:environments}
  \end{center}
\end{figure}

The gym minigrid provides various environments, but in this study, we used three environments that need language understanding to allow the agent to get the reward. 
Fig. \ref{fig:environments} shows the three environments that we used in this study and the details of the language instructions, objects, and reward conditions given in each environment.
In these environments, the visual information obtained by the agent is in form of $7\times7$ squares in front of the agent.
In addition, the same language instruction is given in the episode for each step.
The goal of the agent is to use these two modalities (visual information and language information) to make appropriate action decisions that understand the language and achieve the language instructions.

\subsubsection{Experimental setting}
To verify multitask learning and transfer learning, the agents were trained by the following three settings:
\begin{itemize}
  \item From scratch: Learn from scratch at the GoToObject environment
  \item Multitask learning: Multitask learning at the Fetch environment and GoToDoor environment
  \item Transfer learning: Learning at the GoToObject environment using the agent learned in the above multitask setting
\end{itemize}
In multitask learning, the model was updated after obtaining the same number of samples from the two environments.

\begin{figure}[tb]
	\begin{center}
	   \includegraphics[width=1.0\linewidth]{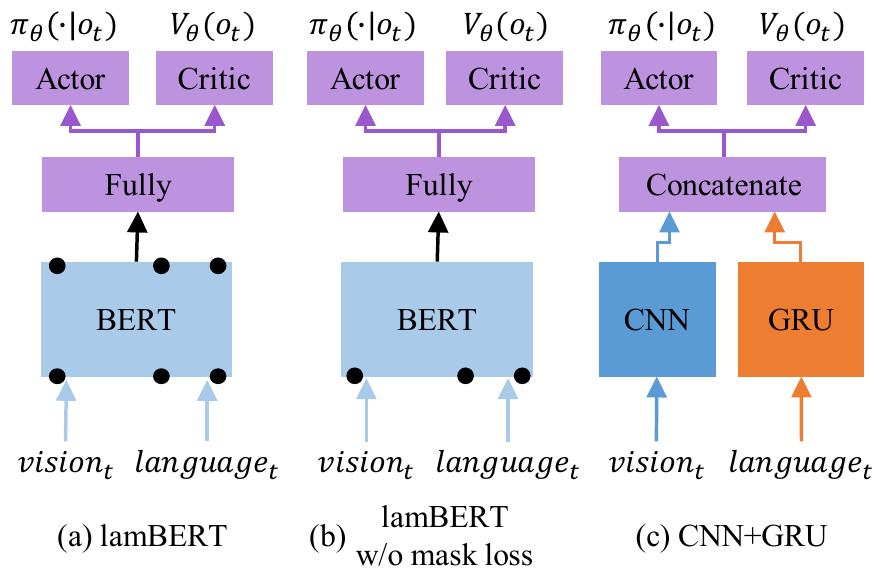}
	   \caption{
	   Comparison models: 
	   (a) lamBERT (prop.), (b) lamBERT (w/o ML), and (c) CNN+GRU. 
	   The hyperparameters of models (a) and (b) are equal.
	   In models (a) and (b), the mask processing is performed on input tokens during learning, but not during action in the environment.
	   The model (b) masks the input tokens during learning but does not update the network using the mask prediction loss $L_{ml}$. 
	   The black dots in the figure represent this situation. 
	   }
	   \label{fig:comparison_models}
   \end{center}
\end{figure}

To compare the proposed methods, three models shown in Fig. \ref{fig:comparison_models} were learned according to the above settings.
As a comparison method, we used the lamBERT model (w/o mask loss (ML)) and a convolutional neural network (CNN)-based model (CNN+GRU) that represents the vision and language information using convolutional neural network (CNN) and gated recurrent unit (GRU), respectively.
When learning the lamBERT (w/o ML), the mask processing was performed on the input token during learning in the same way, as in the lamBERT (w/ ML).
This processing of the lamBERT (w/o ML) was performed to eliminate the possibility that the input of data with a mask by the proposed method could improve the generalization of the model.
The lamBERT model's hyperparameters are shown in Table \ref{table:hyperparameters}.

\begin{table}[tb]
\begin{center}
\caption{
Hyperparameters of lamBERT models.
The parameters are shown separately to the BERT part and the PPO part.
}
\label{table:hyperparameters}
\begin{tabular}{|c|c||c|c|}
\hline
\multicolumn{2}{|c||}{BERT network} & \multicolumn{2}{c|}{PPO network} \\ \hline \hline
embedding token size  & 64 & $c_1$        & 0.5        \\ \hline
num of attention head & 4  & $c_2$        & 0.01       \\ \hline
num of hidden layers  & 3  & $\epsilon$           & 0.2        \\ \hline
\end{tabular}
\end{center}
\end{table}

\subsubsection{Experimental Results}

\begin{figure*}[ht]
  \centering
    \begin{tabular}{cc}
 
      \begin{minipage}{0.33\hsize}
        \centering
          \includegraphics[width=1.0\linewidth]
                          {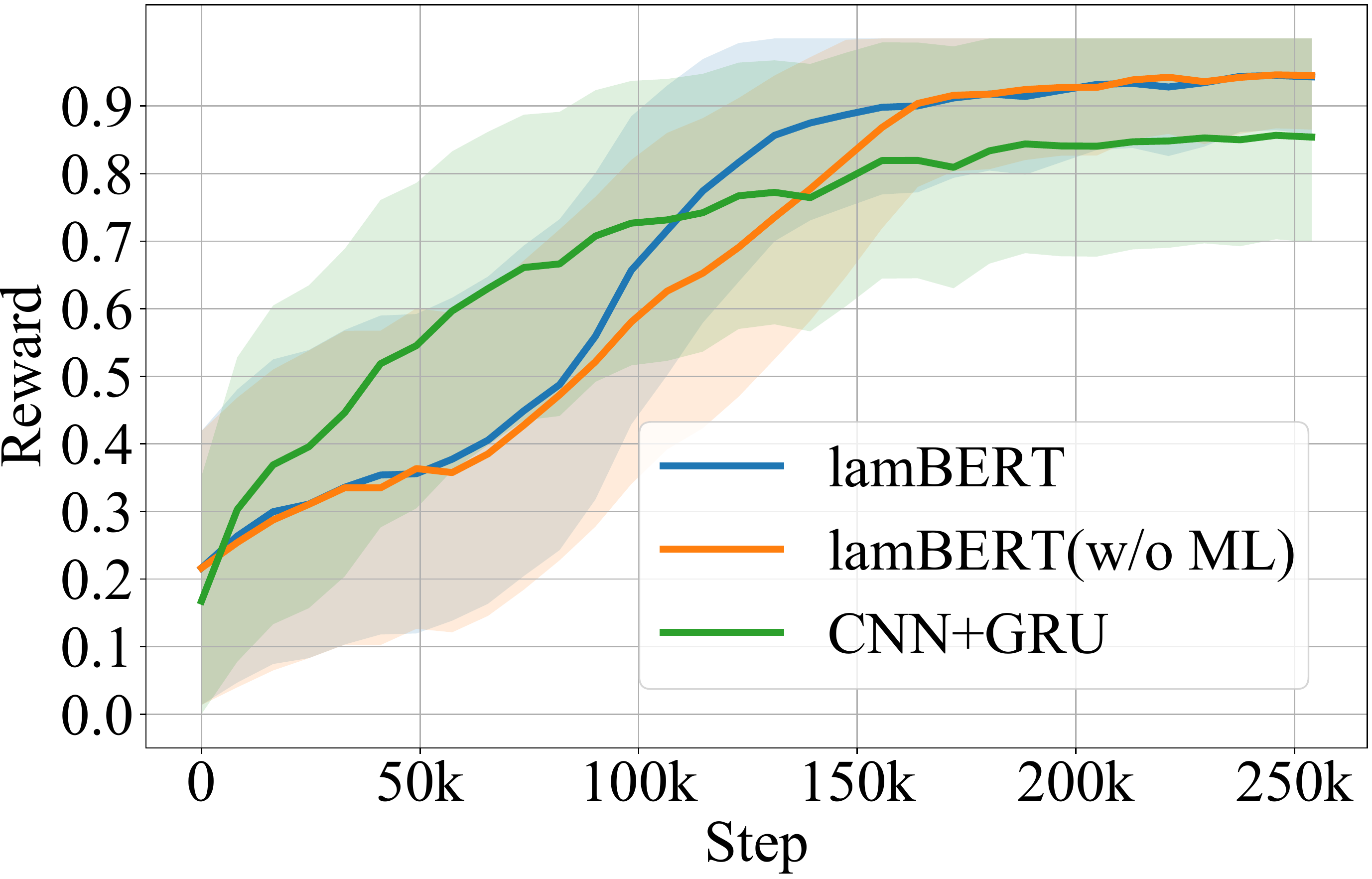}
                          \subcaption{From scratch}
      \end{minipage}
 
      \begin{minipage}{0.33\hsize}
        \centering
          \includegraphics[width=1.0\linewidth]
                          {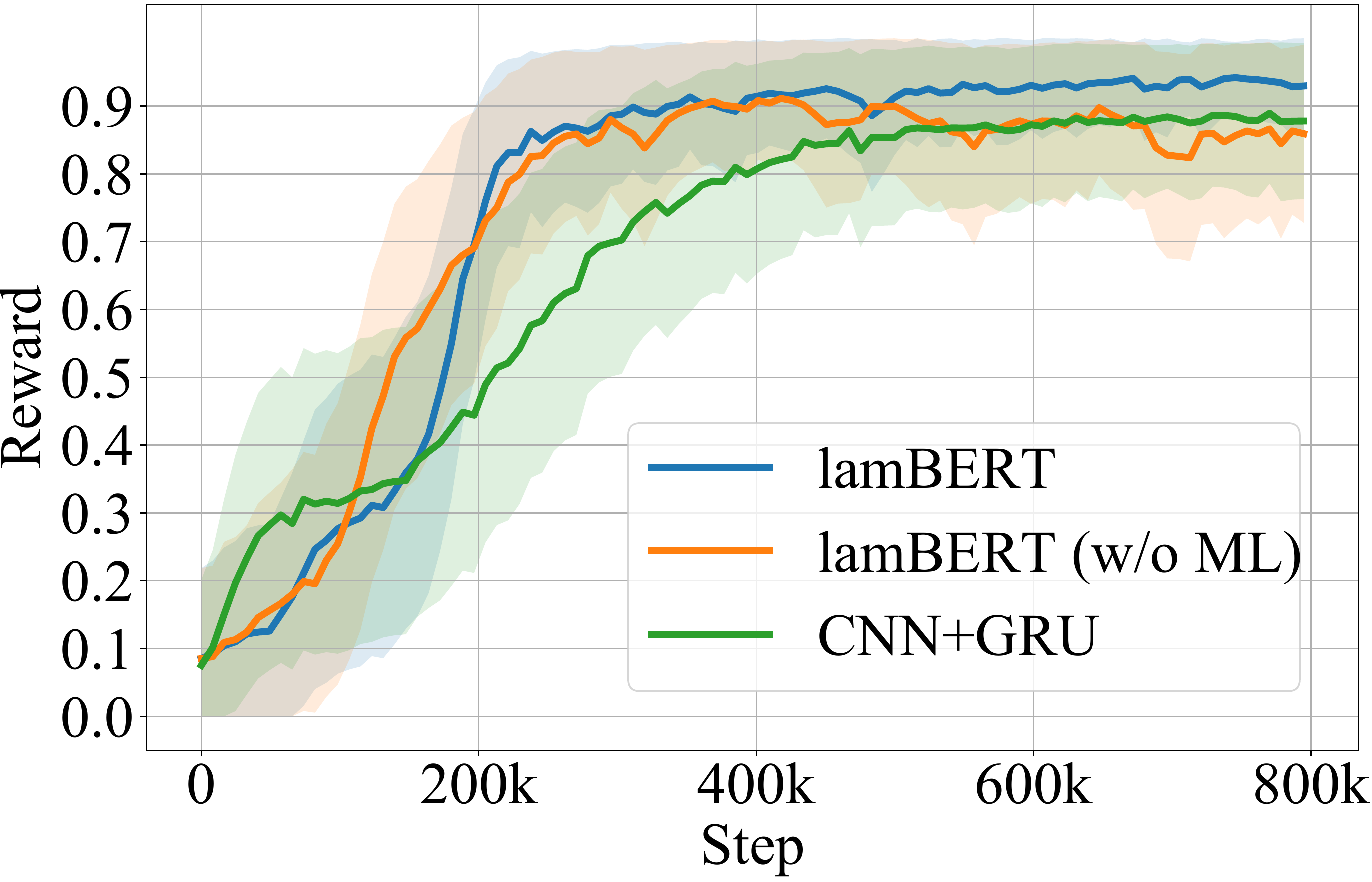}
                          \subcaption{Multitask learning}
      \end{minipage}
 
      \begin{minipage}{0.33\hsize}
        \centering
          \includegraphics[width=1.0\linewidth]
                          {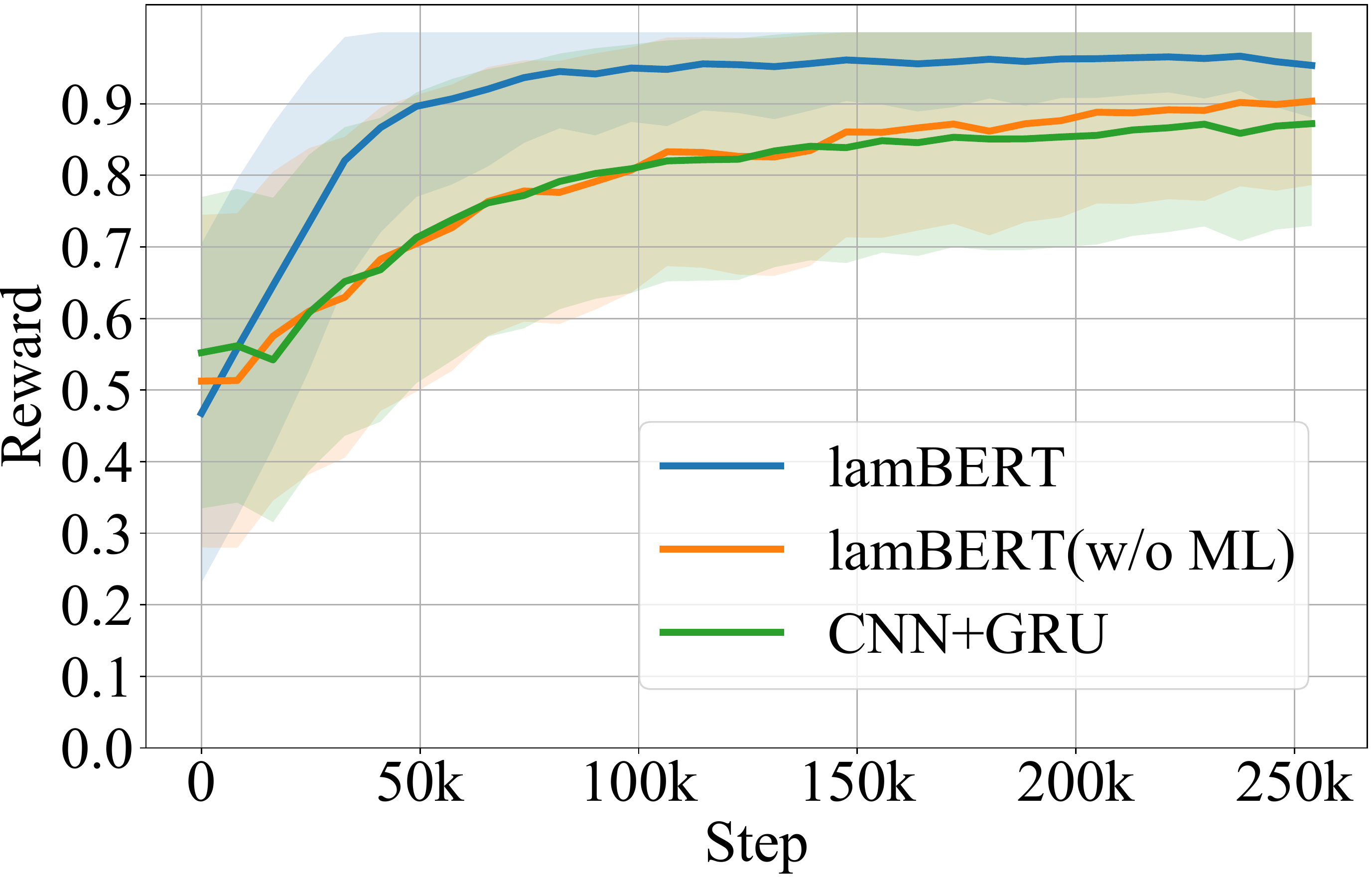}
                          \subcaption{Transfer learning}
                          
      \end{minipage}
    
    \end{tabular}
    \caption{
    The Reward value for each model in each environment; 
    (a) from scratch at GotoObject, (b) multitask learning at Fetch and GotoDoor, and (c) transfer learning at GotoObject: 
    The results of each model are the average and standard deviation ($0.5\sigma$) of reward values when the seed value was changed three times.
    Each model updates its parameters every 8064 steps. 
    In the transfer learning, the last model learned in the multitask setting was used for the initial weight of the transfer setting.
    In the multitask setting, the number of samples obtained from each environment is equal. 
    }
    \label{fig:results_reward}
\end{figure*}

\begin{figure}[tb]
	\begin{center}
	   \includegraphics[width=1.0\linewidth]{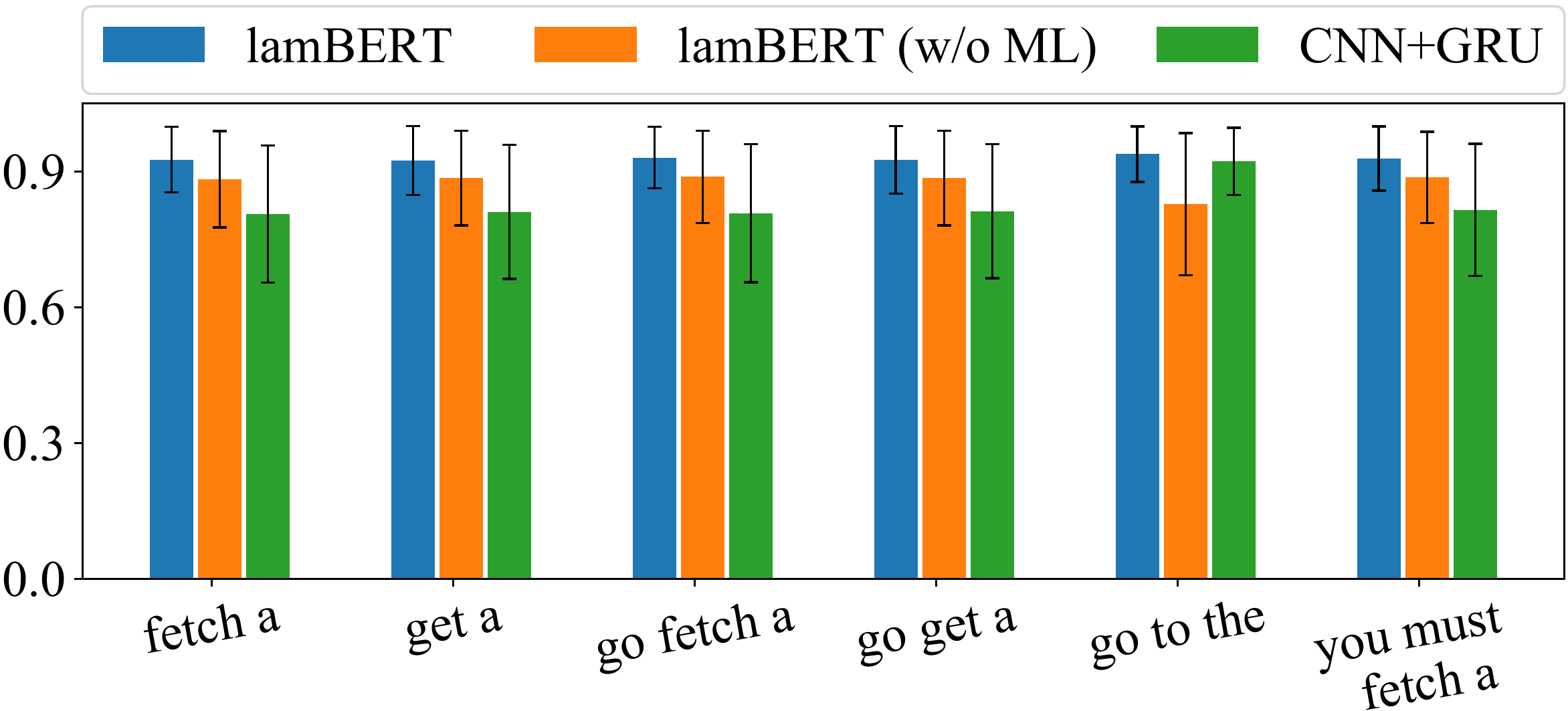}
	   \caption{
	   Reward value for each language instruction: 
	   The average and standard deviation ($0.5$ $\sigma$) of reward values for each language instruction in multitask setting are shown.
	   For each model, the results obtained from the learned model from three seed values are shown. 
	   For each language instruction, the reward values are averaged over the colors and object names. 
	   Hence the reward values for different verbs are given in the figure. 
	   }
	   \label{fig:reward_each_mission}
   \end{center}
\end{figure}

\begin{figure*}[ht]
  \centering
    \begin{tabular}{cc}
 
      \begin{minipage}{1.0\hsize}
        \centering
          \includegraphics[width=1.0\linewidth]
                          {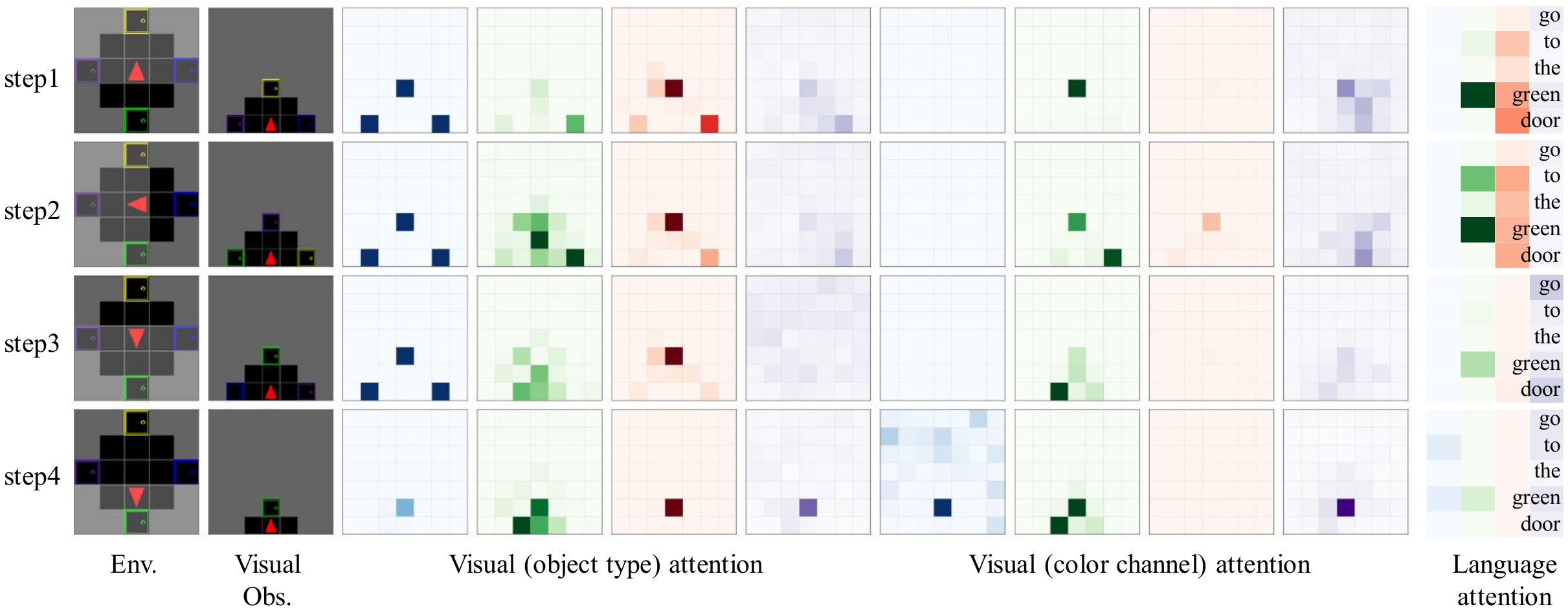}
                          \subcaption{GotoDoor Environment (mission: go to the green door)}
      \end{minipage}\\
 
      \begin{minipage}{1.0\hsize}
        \centering
          \includegraphics[width=1.0\linewidth]
                          {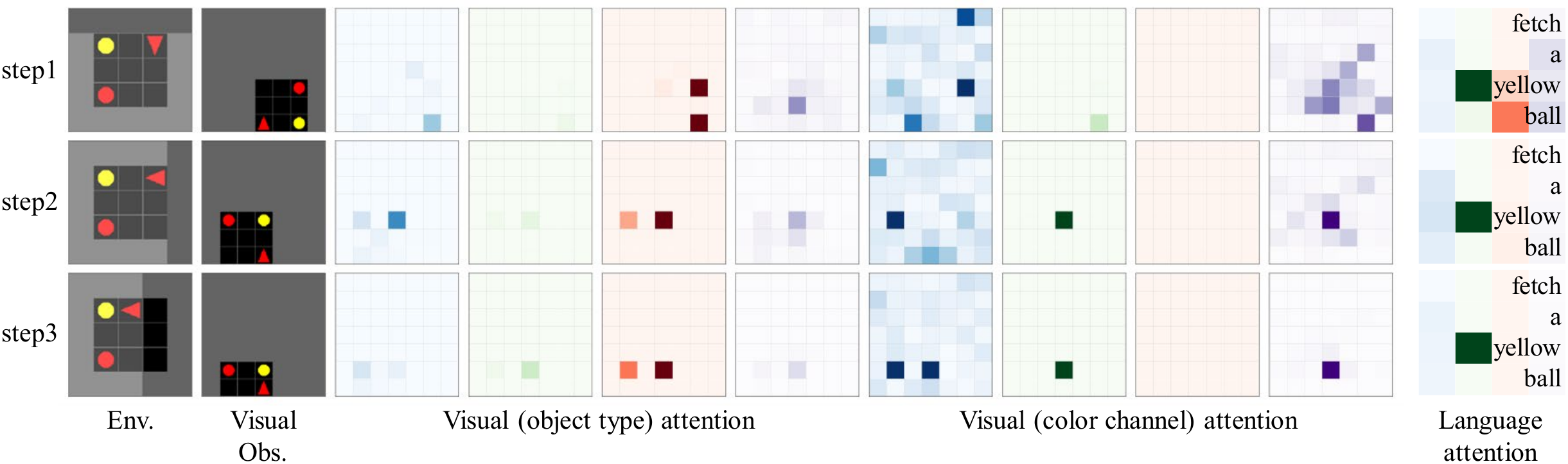}
                          \subcaption{Fetch Environment (mission: fetch a yellow ball)}
      \end{minipage}
 
    \end{tabular}
    \caption{
    Visualization of attention weight: 
    This figure shows the attention weight for the CLS token in each step. 
    The color of each attention indicates the type of attention-head.
    The attention of visual information is shown separately for two channels of visual information (object type and color). 
    The attention for language information indicates the attention weight of each attention-head for each word of the mission. 
    The attention is shown only for the second layer of the lamBERT model. 
    We used an agent model that learned the lamBERT (w/ ML) in the multitask setting to obtain this result. 
    }
    \label{fig:attention}
\end{figure*}

The mean and standard deviation ($0.5$ $\sigma$) of rewards for each model in each setting is shown in Fig. \ref{fig:results_reward}.
We performed each setting and each model learning three times with different random seed values.
First, by looking at the results of Fig. \ref{fig:results_reward} (a) GotoObject, initially, the lamBERT and lamBERT (w/o ML) have lower average rewards than CNN+GRU, but as learning progresses, lamBERT models obtain higher average rewards.

Next, we explain multitask learning.
Fig. \ref{fig:results_reward} (b) is the result of learning at Fetch Environment and GoToDoor Environment at the same time.
This figure shows that the CNN+GRU model obtained a higher reward value in the initial learning, but as the learning progressed, the lamBERT and lamBERT (w/o ML) obtained a higher reward value.
In addition, the results of around 600 - 800 k steps show that the lamBERT obtained the highest average reward value.

Based on these results, we verified the method by which each model was able to take appropriate action decisions for each language instruction in a multitasking environment.
In the verification, based on each model learned in the multitask environment, the agent again acted in the multitask environment and summarized the amount of reward obtained for each language instruction.
The result is shown in Fig. \ref{fig:reward_each_mission}.
This figure shows that the lamBERT received a high reward for all language instructions.
By looking at the lamBERT (w/o ML), the reward value for the language instruction ``go to the" was lower than the reward value for other language instructions.
On the contrary to the lamBERT (w/o ML), the reward value of the CNN+GRU model for the language instruction ``go to the" was high, but the reward value for other language instructions was low.
The language instruction ``go to the" belongs to the GotoDoor Environment, and the other language instructions belong to the Fetch Environment.
In summary, the lamBERT (w/o ML) and the CNN+GRU model could not act properly in the two environments, but the lamBERT model can act properly in all environments.
These results suggest that the lamBERT model's mask prediction task is effective for reinforcement learning tasks in a multitasking environment.

Then, we explain the results of transfer learning.
Fig. \ref{fig:results_reward} (c) shows the result of learning in the GoToObject Environment using a model learned in a multitask setting.
Fig. \ref{fig:results_reward} (c) shows that all models can get higher reward values compared to learning from scratch (Fig. \ref{fig:results_reward} (a)).
Comparing the results of transfer learning, the lamBERT obtained the highest reward values and was the fastest to obtain a high reward value as compared to other methods.
The lamBERT (w/o ML) and CNN+GRU models, which are comparison methods, have obtained almost the same reward values.
These results suggest that the lamBERT model's mask prediction task is effective for transfer learning in reinforcement learning tasks.

Finally, we explain the results of attention weight.
Fig. \ref{fig:attention} shows the attention weight when the agent makes an action decision.
Fig. \ref{fig:attention} (a) shows the attention weight of the agent acting on GotoDoor Environment under the mission ``go to the green door." 
Fig. \ref{fig:attention} (b) shows the attention weight at the agent acting on Fetch Environment under the mission ``fetch a yellow ball." 
The attention weight in Fig. \ref{fig:attention} indicates the attention of the CLS token.
This is because, as shown in Fig. \ref{fig:lamBERT}, the actor-critic network uses the CLS token.
The color of each attention weight indicates the type of the attention head.
These results are obtained by using the lamBERT model learned in the multitask setting.
There are two types of attention, namely, visual attention and language attention.
Visual attention corresponds to partial visual observation and is divided into two types: object type attention and color channel attention as the visual observation has two channels of object type and color.
The language attention indicates the attention weight of each attention head corresponding to each word of the mission.
In Fig. \ref{fig:attention}, the attention is shown only for the second layer of the lamBERT model.

Fig. \ref{fig:attention} shows that the visual (object type) attention tends to take a higher value for a certain position of an object such as a door or a ball.
Further, from the visual (color channel) attention (green attention head) in Fig. \ref{fig:attention} (b), the attention weight is high only for the ``yellow" ball.
Furthermore, the language attention (green attention head) shows a high value for the word for color.
These results indicate that the attention weight for important information is high.
Further, as shown by the green attention head result, it is considered that the model obtained the multimodal attention representation.

%% file: 4discussion.tex
\section{Discussion}
The purpose of the present study is to show that using the structure and learning method of BERT, effective representation for action and language learning of autonomous agents can be learned.
Experimental results show that the proposed method can obtain higher rewards than other models in the multitask and transfer learning settings.
We discuss these results from two aspects: 1) BERT representation learning and 2) reinforcement learning.

\subsubsection{BERT representation learning}
The proposed method is a model in which the reinforcement learning is applied to the fine-tuning task of BERT.
By applying the reinforcement learning to BERT, agents can perform data acquisition and mask prediction tasks simultaneously.
Thus, it is possible to autonomously make the model learn while collecting data to understand action and language and performing a mask prediction task in the current environment, as pre-training for a new environment.
The results (Fig. \ref{fig:attention}) show that the attention is applied to important information for both vision and language information.
From these results, it is considered that the lamBERT model learned multimodal representation, which is important for the action decision.
The results of transfer learning in Fig. \ref{fig:results_reward} (c) demonstrate that the agent with the proposed model adapted faster and gained higher rewards than the other models in the new environment. 
As described above, by performing the pre-training task (mask prediction task) in the pre-transition environment, the agent can better understand the language and the environment for acting in a new environment.
We consider in the proposed method that the learning of representation in a specific environment was acquired by reinforcement learning along with acquiring general representation such as grammatical structures and environmental structures by mask prediction tasks. 
We consider that we can analyze whether such representation learning has been performed by using the attention weight, as shown in Fig. \ref{fig:attention}. 
Furthermore, by comparing the difference in the attention weight between the models with and without the mask prediction task, we think it will be possible to analyze the effect of the mask prediction task on decision making.

\subsubsection{Reinforcement learning}
The mask prediction task in the proposed model can be considered as an auxiliary task, which is often used in reinforcement learning.
The auxiliary task is designed in addition to the main task (cumulative reward maximization) in reinforcement learning (e.g., prediction of reward value or prediction of other modalities). 
As reported, such an auxiliary task has an advantage in obtaining the agents' rewards \cite{jaderberg2016reinforcement, hermann2017grounded}.
The proposed model can be regarded as a reinforcement learning model using a mask prediction task as an auxiliary task.
We consider that this is one reason that the mask prediction task was effective in obtaining the rewards in a multitask environment and transfer learning.
The auxiliary task has a problem in that the effective design of the auxiliary task is unknown.
It is useful for a mask prediction task to be used as an auxiliary task because a specific auxiliary task does not need to be manually designed.
Furthermore, if the relationship between the mask prediction task in the BERT and the auxiliary task in reinforcement learning becomes clear, better mask candidates can be selected.
In addition, from the viewpoint of reinforcement learning, the effects of the auxiliary tasks may be interpreted by analyzing an attention weight of the BERT model.

\subsubsection{Limitations of the proposed model}
Finally, we describe the three limitations.
First, the proposed model cannot have anything other than a token sequence as input.
In other words, the proposed model only handles discrete information, such as language and visual information, in the grid environment.
However, the robots need continuous multimodal information, such as sounds, haptics, and motor commands.
This problem can be addressed by applying the mask prediction task to continuous values using regression.
 
Second, the proposed model makes it difficult to process high-dimensional data because self-attention requires a computational cost of $ O(n^2) $ for an input sequence of length $n$.
Thus, reduction of the calculation cost and number of parameters for self-attention is one solution, to which several methods have been proposed \cite{lan2019albert, kitaev2020reformer}.
We consider that these methods are also applicable to the proposed method.

Third, the proposed model cannot learn the multimodal representation when each modality has a dependency over time because the proposed method is learned using data at a certain time $t$.
However, in general, the modalities have dependencies over time.
For example, the language information observed by the agent is not only closely related to a certain time $t$ but also to a state-action time series.
Therefore, it is important to understand the multimodal relationship between such a state-action time sequence and language information.
One way to obtain this representation is by extending the input of the model to action-state time series information.

%% file: 5conclusion.tex
\section{Conclusion}

In this study, we propose the lamBERT model, in which the robot learns concepts from multiple sensory inputs using the BERT structure, and learns actions using learned multimodal representation and reinforcement learning. 
We verified the usefulness of the model using a grid environment that required to understand the language instructions and showed that the structure and learning method of the BERT is effective for multitask learning and transfer learning. 
In this study, we evaluated the model mainly by looking at the results of the agent actions.
However, in the future, we will analyze the internal representation of the model and the effects of representation learning on decision making.
For another future work, we will extend the model to adapt the proposed model to a real robot.
We consider the integration of continuous and high-dimensional sensory input and motor information with discrete language information.
Furthermore, we aim to learn more complex language environment relationships by extending the model over time.

%% file: 0root.bbl
\begin{thebibliography}{10}

\bibitem{taniguchi2016symbol}
T.~Taniguchi, T.~Nagai, T.~Nakamura, N.~Iwahashi, T.~Ogata, and H.~Asoh,
  ``Symbol emergence in robotics: a survey,'' {\em Advanced Robotics}, vol.~30,
  no.~11-12, pp.~706--728, 2016.

\bibitem{nakamura2009grounding}
T.~Nakamura, T.~Nagai, and N.~Iwahashi, ``Grounding of word meanings in
  multimodal concepts using lda,'' in {\em 2009 IEEE/RSJ International
  Conference on Intelligent Robots and Systems}, pp.~3943--3948, IEEE, 2009.

\bibitem{araki2012online}
T.~Araki, T.~Nakamura, T.~Nagai, S.~Nagasaka, T.~Taniguchi, and N.~Iwahashi,
  ``Online learning of concepts and words using multimodal lda and hierarchical
  pitman-yor language model,'' in {\em 2012 IEEE/RSJ International Conference
  on Intelligent Robots and Systems}, pp.~1623--1630, IEEE, 2012.

\bibitem{araki2013long}
T.~Araki, T.~Nakamura, and T.~Nagai, ``Long-term learning of concept and word
  by robots: Interactive learning framework and preliminary results,'' in {\em
  2013 IEEE/RSJ International Conference on Intelligent Robots and Systems},
  pp.~2280--2287, IEEE, 2013.

\bibitem{suzuki2016joint}
M.~Suzuki, K.~Nakayama, and Y.~Matsuo, ``Joint multimodal learning with deep
  generative models,'' {\em arXiv preprint arXiv:1611.01891}, 2016.

\bibitem{NIPS2019_9702}
Y.~Shi, S.~N, B.~Paige, and P.~Torr, ``Variational mixture-of-experts
  autoencoders for multi-modal deep generative models,'' {\em Advances in
  Neural Information Processing Systems 32}, pp.~15718--15729, 2019.

\bibitem{levine2016end}
S.~Levine, C.~Finn, T.~Darrell, and P.~Abbeel, ``End-to-end training of deep
  visuomotor policies,'' {\em The Journal of Machine Learning Research},
  vol.~17, no.~1, pp.~1334--1373, 2016.

\bibitem{finn2017model}
C.~Finn, P.~Abbeel, and S.~Levine, ``Model-agnostic meta-learning for fast
  adaptation of deep networks,'' in {\em Proceedings of the 34th International
  Conference on Machine Learning-Volume 70}, pp.~1126--1135, JMLR. org, 2017.

\bibitem{dasari2019robonet}
S.~Dasari, F.~Ebert, S.~Tian, S.~Nair, B.~Bucher, K.~Schmeckpeper, S.~Singh,
  S.~Levine, and C.~Finn, ``Robonet: Large-scale multi-robot learning,'' {\em
  arXiv preprint arXiv:1910.11215}, 2019.

\bibitem{peters2018deep}
M.~Peters, M.~Neumann, M.~Iyyer, M.~Gardner, C.~Clark, K.~Lee, and
  L.~Zettlemoyer, ``Deep contextualized word representations,'' in {\em
  Proceedings of the 2018 Conference of the North American Chapter of the
  Association for Computational Linguistics: Human Language Technologies,
  Volume 1 (Long Papers)}, pp.~2227--2237, 2018.

\bibitem{devlin2018bert}
J.~Devlin, M.-W. Chang, K.~Lee, and K.~Toutanova, ``Bert: Pre-training of deep
  bidirectional transformers for language understanding,'' {\em arXiv preprint
  arXiv:1810.04805}, 2018.

\bibitem{sun2019videobert}
C.~Sun, A.~Myers, C.~Vondrick, K.~Murphy, and C.~Schmid, ``Videobert: A joint
  model for video and language representation learning,'' in {\em Proceedings
  of the IEEE International Conference on Computer Vision}, pp.~7464--7473,
  2019.

\bibitem{lu2019vilbert}
J.~Lu, D.~Batra, D.~Parikh, and S.~Lee, ``Vilbert: Pretraining task-agnostic
  visiolinguistic representations for vision-and-language tasks,'' in {\em
  Advances in Neural Information Processing Systems}, pp.~13--23, 2019.

\bibitem{qi2020imagebert}
D.~Qi, L.~Su, J.~Song, E.~Cui, T.~Bharti, and A.~Sacheti, ``Imagebert:
  Cross-modal pre-training with large-scale weak-supervised image-text data,''
  {\em arXiv preprint arXiv:2001.07966}, 2020.

\bibitem{vedantam2017generative}
R.~Vedantam, I.~Fischer, J.~Huang, and K.~Murphy, ``Generative models of
  visually grounded imagination,'' {\em arXiv preprint arXiv:1705.10762}, 2017.

\bibitem{SPCOSLAM2017IROS}
A.~{Taniguchi}, Y.~{Hagiwara}, T.~{Taniguchi}, and T.~{Inamura}, ``Online
  spatial concept and lexical acquisition with simultaneous localization and
  mapping,'' in {\em 2017 IEEE/RSJ International Conference on Intelligent
  Robots and Systems (IROS)}, pp.~811--818, Sep. 2017.

\bibitem{jaderberg2016reinforcement}
M.~Jaderberg, V.~Mnih, W.~M. Czarnecki, T.~Schaul, J.~Z. Leibo, D.~Silver, and
  K.~Kavukcuoglu, ``Reinforcement learning with unsupervised auxiliary tasks,''
  {\em arXiv preprint arXiv:1611.05397}, 2016.

\bibitem{li2018rethinking}
Z.~Li, T.~Motoyoshi, K.~Sasaki, T.~Ogata, and S.~Sugano, ``Rethinking
  self-driving: Multi-task knowledge for better generalization and accident
  explanation ability,'' {\em arXiv preprint arXiv:1809.11100}, 2018.

\bibitem{chebotar2019closing}
Y.~Chebotar, A.~Handa, V.~Makoviychuk, M.~Macklin, J.~Issac, N.~Ratliff, and
  D.~Fox, ``Closing the sim-to-real loop: Adapting simulation randomization
  with real world experience,'' in {\em 2019 International Conference on
  Robotics and Automation (ICRA)}, pp.~8973--8979, IEEE, 2019.

\bibitem{hermann2017grounded}
K.~M. Hermann, F.~Hill, S.~Green, F.~Wang, R.~Faulkner, H.~Soyer,
  D.~Szepesvari, W.~M. Czarnecki, M.~Jaderberg, D.~Teplyashin, {\em et~al.},
  ``Grounded language learning in a simulated 3d world,'' {\em arXiv preprint
  arXiv:1706.06551}, 2017.

\bibitem{miyazawa2019integrated}
K.~Miyazawa, T.~Horii, T.~Aoki, and T.~Nagai, ``Integrated cognitive
  architecture for robot learning of action and language,'' {\em Frontiers in
  Robotics and AI}, vol.~6, p.~131, 2019.

\bibitem{zhu2019vision}
F.~Zhu, Y.~Zhu, X.~Chang, and X.~Liang, ``Vision-language navigation with
  self-supervised auxiliary reasoning tasks,'' {\em arXiv preprint
  arXiv:1911.07883}, 2019.

\bibitem{kiela2019supervised}
D.~Kiela, S.~Bhooshan, H.~Firooz, and D.~Testuggine, ``Supervised multimodal
  bitransformers for classifying images and text,'' {\em arXiv preprint
  arXiv:1909.02950}, 2019.

\bibitem{schulman2017proximal}
J.~Schulman, F.~Wolski, P.~Dhariwal, A.~Radford, and O.~Klimov, ``Proximal
  policy optimization algorithms,'' {\em arXiv preprint arXiv:1707.06347},
  2017.

\bibitem{chevalier-boisvert2018babyai}
M.~Chevalier-Boisvert, D.~Bahdanau, S.~Lahlou, L.~Willems, C.~Saharia, T.~H.
  Nguyen, and Y.~Bengio, ``Baby{AI}: First steps towards grounded language
  learning with a human in the loop,'' in {\em International Conference on
  Learning Representations}, 2019.

\bibitem{lan2019albert}
Z.~Lan, M.~Chen, S.~Goodman, K.~Gimpel, P.~Sharma, and R.~Soricut, ``Albert: A
  lite bert for self-supervised learning of language representations,'' {\em
  arXiv preprint arXiv:1909.11942}, 2019.

\bibitem{kitaev2020reformer}
N.~Kitaev, {\L}.~Kaiser, and A.~Levskaya, ``Reformer: The efficient
  transformer,'' {\em arXiv preprint arXiv:2001.04451}, 2020.

\end{thebibliography}
